\documentclass[letterpaper]{article} 
\usepackage{aaai25}
\usepackage{times}  
\usepackage{helvet}  
\usepackage{courier}  
\usepackage[hyphens]{url}  
\usepackage{graphicx} 
\usepackage{natbib}  
\usepackage{caption} 
\usepackage{multirow}
\usepackage{booktabs}
\usepackage{amsmath}
\usepackage[ruled,linesnumbered,vlined,noend]{algorithm2e}
\usepackage{subfigure}

 	\newcommand{\presec}{\vspace{0cm}}
	\newcommand{\postsec}{\vspace{0cm}}
	\newcommand{\presub}{\vspace{0cm}}
	\newcommand{\postsub}{\vspace{0cm}}

    \newcommand{\bbb}{\noindent\textbf}

\title{An Efficient Inference Framework for Early-exit Large Language Models}
\author{
    Ruijie Miao,
    Yihan Yan,
    Xinshuo Yao,
    Tong Yang
}
\affiliations {
    Peking University
}

\begin{document}

\maketitle

\begin{abstract}
    Building efficient inference framework has gained increasing interests for research community.
    Early-exit models, a variant of LLMs, improves the inference efficiency of LLMs by skipping rest layers and directly generate output tokens when they are confident enough.
    However, there is no work of LLM inference framework that takes early-exit models into consideration.
    This is non-trivial as prior art on LLM inference cannot be directly applied to early-exit models.
    In this work, we solves two key challenges in building efficient inference framework for early-exit models: (1) batch inference at iteration-level granularity; and (2) KV cache management.
    For the former, we propose to process the batch until all sequences surpass the early-exit confidence threshold.
    For the latter, we propose to fill the KV cache of rest layers before the iteration terminates.
    Our evaluation shows that, compared with the original vLLM operating at full layers, our solution achieves up to $1.25\times$ speed up.
\end{abstract}

%

 	\presec
\section{Introduction}
\label{sec:intro}
\postsec


In recent years, various Large Language Models (LLMs) \cite{gpt-4, llama, palm, Chinchilla} have achieved excellent performance in different tasks, including translation \cite{prompt-translation, multilingual-translation-survey}, classification \cite{unintended-bias-classification, few-shot-text-classification}, question answering \cite{QA-Unifiedqa, natural-questions, openbookQA}, and more \cite{vx2text, news-summary}.
The wide applications of LLMs in real-world scenarios lead to an increased demand for the inference services.
Researchers have paid much attention to optimize the inference framework, such as FasterTransformer \cite{FasterTransformer}, Orca \cite{orca}, vLLM \cite{vllm}.
The goal is to improve the serving throughput and efficiently manage the GPU memory.

Early-exit models \cite{early-exit-bert, hash-early-exit, depth-adapt-transforemr, CATs, calm} have been proposed as a promising variant of LLMs.
Normal LLMs achieves good performance with huge model size and large number of model layers, which consumes much computation resources during the inference.
Early-exit models observes that such large number of layers is often unnecessary and ``skipping" some layers during the inference can result in comparable performance.
Specifically, during the inference of each token, every layer will determine whether the current output is sufficiently accurate. If it is, the remaining layers can be skipped to directly obtain the output token.
By skipping layers, early-exit models can reduce the computation load  of the decoding process, and therefore the serving throughput can be improved.
As the inference speed of LLMs gains increasing attention, the advantages of early-exit models in enhancing inference speed are being increasingly recognized.
For example, Adainf \cite{adainf} applies the early-exit technique to accelerate the inference.
We believe the early-exit models can be promising alternatives, especially for LLM inference.

In this work, we aim to build efficient inference framework for early-exit LLMs.
Prior work on LLM inference framework is designed for common LLMs, and as far as we know, there is no work to optimize the inference framework of the early-exit LLMs.
Actually those work that proposes early-exit models usually tests them with batch size equal to $1$.
Directly applying the inference framework of normal LLMs to the early-exit LLMs is inefficient due to following reasons.
First, the early-exit LLMs explicit different inference process, and therefore brings new opportunity in improving the inference efficiency.
Prior inference frameworks on the normal LLMs do not take the early-exit techniques into considerations, and therefore can miss the opportunity to skip the extra computation overhead.
Besides, the skipping behaviour brings challenges in the KV cache management, as the KV tensors are not computed for skipped layers.
Building inference framework for early-exit LLMs also requires re-designing of the KV cache management module.

We propose a unified framework for accelerating the inference of early-exit LLMs, supporting different early-exit techniques.
Our solution is based on the prior art of iteration-level scheduling \cite{orca}, which batches the request sequences and breaks the decoding procedure into multiple iterations, each iteration decoding one token for each sequence in the batch.
To support early-exit LLMs, for each iteration we record whether the current request sequences in the batch can exit early.
Every layer we update the early-exit status for those request sequences that are determined to be eligible for early exit.
Once all sequences in the batch are marked as eligible for early exit, we terminate the current iteration and output tokens for all sequences in the batch.
To support KV cache reuse for skipped layers, before the termination of each iteration, we further calculate the KV cache for skipped layers.
Therefore, our work successfully adapts existing state-of-the-art inference frameworks to early-exit models.

We implement our solution on vLLM \cite{vllm} in python.
Although there are many open-source large models, there are very few open-source early-exit models.
Therefore, we implement the recently proposed CALM \cite{calm}, which is an early-exit version of the T5 model.
We implement three popular early-exit techniques: softmax response, hidden states similarity, and dedicated early-exit classifier.
We perform experiments on CALM with these three early-exit techniques to see their performance on inference.
Our results show that the early-exit technique of hidden states similarity reaches highest early exit rates and best inference speed.
Our framework achieves up to $1.2\times$ speed up compared with vLLM on full layers.

 	\presec
\section{Background and Motivation}
\label{sec:background}
\postsec

In this section, we first introduce the early-exit strategies in the neural network models in \S \ref{sec:background:early-exit}. Then we discuss current works on the inference systems for transformer-based models in \S \ref{sec:background:inference}.

\presub
\subsection{Early-exit Models}
\label{sec:background:early-exit}
\postsub



Accelerating the inference of the neural network models is an important topic in machine learning.
Early-exit model is a popular method for speeding up inference.
When generating a token, early-exit models may stop at early layers and directly generate a token without executing the rest of layers.
Compared with normal models, early-exit models execute much less number of layers in average.
The key problem is how to decide whether the model should early exit at one layer.
Intuitively, if the model is ``confident'' that at the current layer they can generate good enough results, they can skip the rest layers.
Below we discuss several ideas for building confidence measurement for early-exit models.
\begin{itemize}
    \item \bbb{Softmax Response:}
    The idea is to compute the difference of the top two candidates as the measurement of confidence. 
    If the gap is large, then the model is confident that it should use the top-1 candidate. 
    Suppose the model uses a matrix $W$ and Softmax operator after all layers as the classifier, and in the $i^{th}$ layer the hidden state is $h_i$ and the confidence threshold is $\lambda_i$. 
    Then We compute,
    $$\text{Top-1}(Softmax(Wh_i)) - \text{Top-2}(Softmax(Wh_i)) > \lambda_i$$
    to decide whether to exit early at the $i^{th}$ layer.

    \item \bbb{Hidden states similarity:}
    the idea is to compute the similarity of two neighboring hidden states as the measurement of confidence. 
    If the similarity of two hidden states is high, we can infer that for the rest layers the resulting hidden states will also be similar \cite{transformer-feed-forward}, and thus we can exit early at the current layer.
    Suppose in the ${i-1}^{th}$ layer and the ${i}^{th}$ layer the hidden states are $h_{i-1}, h_i$, respectively, the confidence threshold is $\lambda_i$ and cosine similarity is used. 
    Then we compute,
    $$cos(h_{i-1}, h_i) > \lambda_i$$
    to decide whether to exit early at the $i^{th}$ layer.

    \item \bbb{Dedicated early-exit classifier:}
    the idea is to introduce a dedicated classifier $M$ to decide whether to exit early.
    Suppose in the ${i}^{th}$ layer the hidden state is $h_i$, and the confidence threshold is $\lambda_i$.
    Then we compute,
    $$M(h_i) > \lambda$$
    to decide whether to exit at the $i^{th}$ layer.

\end{itemize}

\bbb{Existing works on early-exit transformer-based models}
The idea of early-exit can be applied to wide ranges of transformer-based models.
CALM \cite{calm} applies the above three ideas on the T5 model \cite{t5} to build early-exit language models.
MuE \cite{MuE} applies the idea of hidden states similarity on the OFA model \cite{OFA} to build early-exit vision language models.
Their code has been merged to the official OFA repository.
Adainf \cite{adainf} applies the early-exit models to accelerate the inference at the edge servers.

\presub
\subsection{Inference Systems for Transformer-based Models}
\label{sec:background:inference}
\postsub

Recently, researchers have made significant progress in building efficient inference systems for transformer-based models.
Two representative works are Orca \cite{orca} and vLLM \cite{vllm}.
Orca studies how to build an efficient inference system for inference on batches of sequences.
Orca breaks the serving of a sequence batch into multiple iterations, each involving the generation of one token.
During the serving of the batch, some sequences may finish early while others are still generating tokens. 
Orca applies the key idea of iteration-level scheduling, which evicts finished sequences and replenishes new ones after each iteration of token generation. 
Orca further explores how to utilize the parallelism within batch inference to speed up the inference.

vLLM points out that the GPU memory becomes a bottleneck for inference, of which the KV cache consumes an large part. 
For transformer-based models, KV cache is a common technique for accelerating the inference at the cost of more GPU memory consumption.
The KV cache stores key and value pairs generated at each iteration for every layer, which can be reused at the attention computation of the same layer for the following token generation.
vLLM finds that pre-allocating GPU memory for KV cache of the maximum length of sequences is wasting, as the sequences may finish early. 
They propose a virtual KV cache management system, which utilizes the idea of virtual memory and dynamically allocate KV cache memory when it is necessary.

Our proposed solution borrows the optimization ideas from previous approaches, specifically including: (1) batch inference, (2) iteration-level scheduling, and (3) efficient KV cache manangement.

 	\presec
\section{System Design}
\label{sec:system}
\postsec

In this section we introduce the design of our inference framework.


\presub
\subsection{Batch Inference at Iteration-level Granularity}
\postsub
\label{sec:system:token-level}

Prior works on the LLM inference show the benefits of running inference for a batch of sequences in a iteration-level granularity.
We aim to achieve batch inference at iteration-level granularity for the early-exit LLMs.
On each iteration of token generation, our framework additionally maintains the early-exit status for each sequence in the batch.
Initially, all sequences are marked as eligible for early-exit.
After each layer of computation, according to the applied early-exit techniques of the early-exit models, each sequence will update its early-exit status.
Specifically, if the early-exit status of one sequence is false, and at the current layer it successfully surpasses the early-exit confidence, then the early-exit status is updated to true.
Once all the sequences' early-exit status are true, then the batch is considered for early exit.
It should be noticed that if the early-exit status of one sequence is already true but the current layer determines that it fails to surpass the early-exit confidence, we still consider the early-exit status to be true.
We believe this is due to some fluctuations caused by the early exit mechanism.
Generally the more layers the inference go through, the more confident they will be to their generated tokens.

\begin{algorithm}[t]
    \renewcommand\baselinestretch{1.1}\selectfont
	\caption{Iteration-level Batch Inference for Early-exit LLMs}
	\label{alg:pseudo:token-level}
	\KwIn{the input tokens $input\_tokens$; the number of layers $n$; the embedding layer $Embed$; the early-exit model $\mathcal{M}$; KV cache $KVCache$; 
	}

        \tcp{Initialization} 
	$hidden\_states \leftarrow Embed(input\_tokens) $; \\
        $Status \leftarrow \text{tensor of length}~n~\text{initialized to}~False$; \\
        \tcp{Batch inference} 
	\For{i = 1 to $n$}{
	    $hidden\_states, accepted \leftarrow \mathcal{M}(i, hidden\_states, KVCache)$; \\
            $Status \leftarrow Status~|~ accepted$; \\
	    \If{torch.all(Status)}{
                $output\_layer \leftarrow i$; \\
	        Break \\
	    }
	}

        \tcp{KV cache filling}
        \For{i = $output\_layer +1$ to $n$} {
            $K, V \leftarrow \mathcal{M}.compute\_KV\_pair(i, hidden\_states)$ \\
            $KVCache.cache(i, K, V)$
        }

        $output\_tokens \leftarrow transform\_to\_tokens(hidden\_states)$; \\
        \KwRet{$output\_tokens$}
\end{algorithm}

\presub
\subsection{KV Cache Management}
\postsub
\label{sec:system:kvcache}

The prior work on KV cache management, VLLM as the representative, targets at the normal LLMs and cannot be directly applied at the early-exit LLMs.
This is because for the normal LLMs, after one iteration, the key and value pairs of all layers at the position are generated and cached.
However, in early-exit LLMs, the key and value pairs after the early-exit layer are not calculated.
Therefore, if the generation of one token requires computation of higher layers while one previous generated token early exits at a lower layer, then the key and value pairs of that token are missed.
However, if we follow the normal LLM inference process, we should go through the rest layers to fill the KV cache, which makes early exit meaningless.
Fortunately, the prior works \cite{depth-adapt-transforemr, calm} point out that, when one iteration early exits at the lower layer, the generated final hidden states can be saturated to the higher layers.
Therefore, our strategy is as follows:
when executing one iteration, we first follow normal computation layer by layer, calculate key and value pairs and cache them in the KV cache in the process.
Then, if the batch decide to early exit at one layer, before the termination of the current iteration, we pass the output hidden states of the last for the KV cache filling for the rest layers.
We use the output hidden state to calculate the key and value pairs of the following layers and store them in the KV cache, which only involves one matrix multiplication operations per layer.
In our strategy, the early-exit layer requires additional computation for filling KV cache of the rest layers.
The final pseudo code of the inference process is shown in Algorithm \ref{alg:pseudo:token-level}.

    
 	\presec
\section{Implementation}
\label{sec:impl}
\postsec

\presub
\subsection{Inference Framework}
\postsub
\label{sec:impl:vllm}

We implement a system prototype of our proposed solution based on vLLM.
The original vLLM is fully designed for decoder-only models, while our target early-exit model, CALM, is finetuned on the encoder-decoder model T5.
Therefore, our implementation is based on the pull request that implements encoder-decoder architecture\footnote{\url{https://github.com/vllm-project/vllm/pull/3117/}}.
We verify our implementation remains the same results for inference compared with standard implementation of the Huggingface transformers library.

\presub
\subsection{Training Details of Early-exit Models}
\postsub
\label{sec:impl:calm}

To evaluate our proposed early-exit inference acceleration framework, we reproduced the early-exit models described in \cite{calm} using T5 v1.1 small (8 layers) and base (12 layers) as the backbone architectures. Specifically:

\begin{itemize}
    \item We implemented the early-exit mechanism for the decoder, allowing exits after each transformer layer in the decoding process.
    
    \item We used the same three confidence measures as the original work: softmax response, hidden-state similarity, and early-exit classifier.
    
    \item For the early-exit classifier, we followed the independent training objective described in the paper, which outperformed the geometric-like objective.
    
    \item We implemented the decaying threshold function to control the trade-off between performance and efficiency.
    
    \item We used PyTorch and the Hugging Face Transformers library for implementation, fine-tuning, and inference, to ensure broader applicability of our approach.
\end{itemize}

We conducted full fine-tuning of both T5 v1.1 small and base models on the CNN/DM dataset. For the T5 v1.1 small model, we trained for approximately 40,000 steps with a batch size of 16. The T5 v1.1 base model was trained for about 300,000 steps using two NVIDIA RTX 4090 GPUs with a batch size of 4. These training configurations were chosen to balance computational resources and model performance.

We validated our implementation by reproducing the performance-efficiency trade-offs reported in the CALM paper for the text summarization task. The results of our reproduction are presented in Table \ref{tab:early-exit-t5-results}, which demonstrates the effectiveness of our implementation in achieving performance-efficiency trade-offs comparable to the original work.

\begin{table*}[htbp]
\centering
\caption{Evaluation of Early-Exit Language Models}
\label{tab:early-exit-t5-results}

\begin{tabular}{lccccc}
\toprule
\multirow{2}{*}{Model} & Early-exit & Calibrated & \multicolumn{2}{c}{ROUGE-L} & Early-exit \\
\cmidrule(lr){3-4}
 & Technique & Threshold & To Full & To Label & Rate (\%) \\
\midrule
\multirow{4}{*}{T5-v1.1-small} & full & -  & - & 0.3524 & - \\
& static-half \footnotemark & - & 0.6575 & 0.3437 & 50.00 \\
& softmax & 0.85 & 0.7670 & 0.3500 & 53.82 \\
& classifier & 0.9  & 0.6731 & 0.3432 & 66.50 \\
& state & 0.95  & 0.6284 & 0.3400 & 56.53 \\
\midrule
\multirow{4}{*}{T5-v1.1-base} & full & -  & - & 0.3603 & - \\
& static-half & - & 0.6352 & 0.3551 & 50.00 \\
& softmax & 0.85 & 0.7493 & 0.3557 & 53.09 \\
& classifier & 0.92  & 0.6868 & 0.3514 & 59.06 \\
& state & 0.96  & 0.6130 & 0.3516 & 57.26 \\
\bottomrule
\end{tabular}
\end{table*}

\footnotetext{The static-half setting is equivalent to the static setting in the original paper, where 'half' indicates the use of only half of the decoder layers. For the small model, this means using 4 out of 8 layers, and for the base model, 6 out of 12 layers.}
 	\presec
\section{Evaluation}
\label{sec:eval}
\postsec

In this section, we evaluate the inference efficiency of our proposed framework by comparing it with full-layers inference on the original vLLM.
Our experiments are done on a server with dual 26-core CPUs (104 threads, Intel(R) Xeon(R) Gold 6230R CPU @ 2.10GHz) and 1 TB memory.
The server is equipped with $6$ NVIDIA RTX 4090 GPUs.
For the experiment, we only use one GPU to evaluate the inference performance on single GPU.

\bbb{Datasets:}
The evaluation is conducted for the news summary task on the CNN/DM \cite{cnndm} dataset, which contains a collection of news articles and corresponding summaries.
The dataset is split to three parts: the training dataset, the validation dataset and the test dataset, which contains $287k$, $13.4k$, $11.5k$ rows separately. 
We use the test dataset for evaluation, where we select rows with the length of news less than $1024$, and uses the first $512$ tokens.

\presub
\subsection{Token Generation Throughput}
\label{sec:eval:thr}
\postsub

\begin{figure*}[!t]
    \centering
    \subfigure[T5-v1.1-small]{
        \includegraphics[width=0.45\textwidth,]{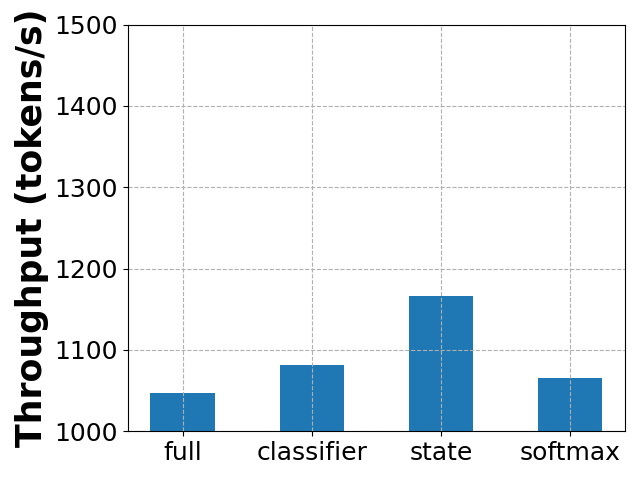}
        \label{fig:thr:small}
    }
    \subfigure[T5-v1.1-base]{
        \includegraphics[width=0.45\textwidth,]{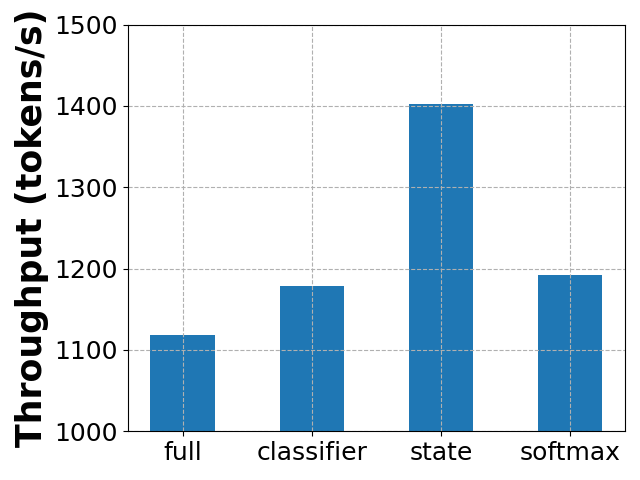}
        \label{fig:thr:base}
    }

    \caption{
    Token generation throughput of two models.
    }
    \label{fig:exp:thr}
\end{figure*}

We compare the token generation throughput of our solution with the original full-layers vLLM on both the T5-v1.1-small and T5-v1.1-base models.
Token generation throughput is defined as the total number of tokens generated divided by the total time of the inference.
The results are shown in Figure \ref{fig:exp:thr}.

For the T5-V1.1-small model, 
the original vLLM running full-layers inference generates tokens at the speed of $1046.47$ tokens/s.
Our proposed solution achieves token generation speed of $1081.04$, $1165.73$, $1065.13$ tokens/s for three different early-exit techniques, respectively.
For the T5-V1.1-base model, 
the original vLLM running full-layers inference generates tokens at the speed of $1117.96$ tokens/s.
Our proposed solution achieves token generation speed of $1777.96$, $1402.78$, $1191.98$ tokens/s for three different early-exit techniques, respectively.

In summary, our proposed solution consistently achieves higher throughput compared with the original vLLM for different early-exit mechanisms.
Besides, using hidden states similarity between layers show the best performance.
Our solution shows up to $1.25\times$ improvement in the token generation throughput.

\presub
\subsection{Inner-token Latency}
\label{sec:eval:latency}
\postsub

\begin{figure*}[!t]
    \centering
    \subfigure[T5-v1.1-small]{
        \includegraphics[width=0.45\textwidth,]{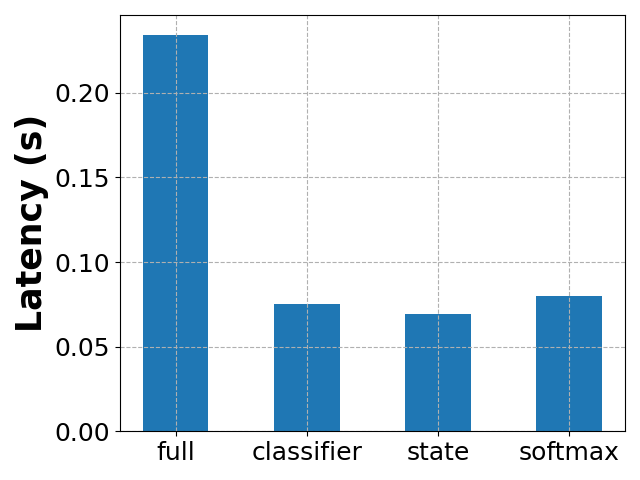}
        \label{fig:latency:small}
    }
    \subfigure[T5-v1.1-base]{
        \includegraphics[width=0.45\textwidth,]{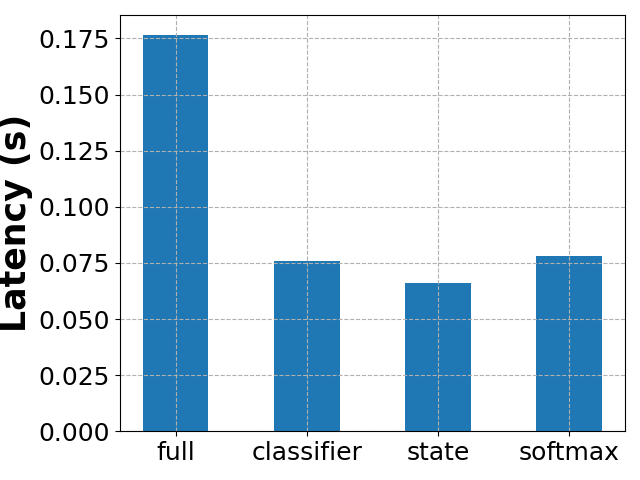}
        \label{fig:lantecy:base}
    }

    \caption{
    Inner-token latency of two models.
    }
    \label{fig:exp:latency}
\end{figure*}

We also compare the inner-token latency of our solution with the original full-layers vLLM on both two models.
For each sequence, we calculate the difference between its finish time and the first-token time as its decoding latency.
The inner-token latency is defined as the sum of decoding latency for all sequences divided by the total number of generated tokens.
The results are shown in Figure \ref{fig:exp:latency}.

For the T5-V1.1-small model, 
the original vLLM running full-layers has inner-token latency of $0.234$s.
Our proposed solution achieves inner-token latency of $0.075$, $0.069$, $0.079$s for three different early-exit techniques, respectively.
For the T5-V1.1-base model, 
the original vLLM running full-layers has inner-token latency of $0.176$s.
Our proposed solution achieves inner-token latency of $0.076$, $0.066$, $0.078$s for three different early-exit techniques, respectively.

In summary, our proposed solution consistently achieves lower inner-token latency compared with the original vLLM for different early-exit mechanisms.
Our solution shows up to $3.39\times$ reduction in the inner-token latency.

\section{Conclusion}


Building efficient inference framework has gained increasing interests for research community.
In this work, we aim to build efficient inference framework for early-exit models, which are variants of LLMs that skip rest decoding layers when confident enough.
Our proposed framework considers the following two aspects.
(1) batch inference at iteration-level granularity; and (2) KV cache management.
For the former, we process the batch until all sequences surpass the early-exit confidence threshold.
For the latter, we fill the KV cache of rest layers before the iteration terminates.
We reproduce and train the state-of-the-art early-exit model CALM and evaluate our inference framework.
Our evaluation shows that, compared with the original vLLM operating at full layers, our solution achieves up to $1.25\times$ speed up.
The results also show the hidden states similarity may be the best mechanism to implement early-exit models when only considering the inference efficiency.

        \clearpage
        \bibliography{main}

\end{document}